\documentclass[runningheads]{llncs}
\usepackage{graphicx}
\graphicspath{{Figures/}}
\usepackage[colorlinks,citecolor=green,linkcolor=red]{hyperref}
\usepackage{amsmath}
\usepackage{booktabs}
\usepackage[misc,geometry]{ifsym} 
\begin{document}
\title{DensSiam: End-to-End  Densely-Siamese Network with Self-Attention Model for Object Tracking}
%
%
%
%
%

\author{Mohamed H. Abdelpakey  \inst{1} $^{(\textrm{\Letter})}$ \and
Mohamed S. Shehata   \inst{1} \and
Mostafa M. Mohamed\inst{2}}

\authorrunning{M. H. Abdelpakey et al.}
\institute{Memorial University of Newfoundland,\\ St. John's, NL A1B 3X5, Canada\\
\email{mha241@mun.ca}\\
 \and
Electrical and Computer Engineering Department,\\ University of Calgary, Canada\\
Biomedical Engineering Department, \\Helwan University, Egypt}

\maketitle              
\begin{abstract}

Convolutional Siamese neural networks have been recently used to track objects using deep features. Siamese architecture can achieve real time speed, however it is still difficult to find a Siamese architecture that maintains the generalization capability, high accuracy and speed while decreasing the number of shared parameters especially when it is very deep. Furthermore, a conventional Siamese architecture usually processes one local neighborhood at a time, which makes the appearance model local and non-robust to appearance changes. 

To overcome these two problems, this paper proposes DensSiam, a novel convolutional Siamese architecture, which uses the concept of dense layers and connects each dense layer to all layers in a feed-forward fashion with a similarity-learning function.  DensSiam also includes a Self-Attention mechanism to force the network to pay more attention to the non-local features during offline training. Extensive experiments are performed on four tracking benchmarks: OTB2013 and OTB2015 for validation set; and VOT2015, VOT2016 and VOT2017 for testing set. The obtained results show that DensSiam achieves superior results on these benchmarks compared to other current state-of-the-art methods.

\end{abstract}

\section{Introduction}

Visual object tracking is an important task in many computer vision applications such as image understanding \cite{lenc15understanding}, surveillance \cite{kendall2015posenet}, human-computer interactions \cite{molchanov2016online} and autonomous driving \cite{chen2015deepdriving}. One of the main challenges in object tracking is how to represent the appearance model in such a way that the model is robust to appearance changes such as motion blur, occlusions, background clutter \cite{wu2015object,alahari2015thermal}. Many trackers use handcrafted features such as  CACF \cite{mueller2017context}, SRDCF \cite {danelljan2015learning}, KCF \cite{henriques2015high} and SAMF \cite{li2014scale} which have inferior accuracy and/or robustness compared to deep features.   

In recent years, deep convolutional neural networks (CNNs) have shown superior performance in various vision tasks. They also increased the performance of object tracking methods. Many trackers have been developed using the strength of CNN features and significantly improved their accuracy and robustness. Deep trackers include SiamFC \cite{bertinetto2016fully}, CFNet \cite{valmadre2017end}, DeepSRDCF \cite{danelljan2015convolutional}, HCF \cite{ma2015hierarchical}. However, these trackers exploit the deep features which originally designed for object recognition and neither consider the temporal information such as SiamFC \cite{bertinetto2016fully} nor non-local features such as the rest of the trackers. The key to design a high-performance tracker is to find the best deep architecture that captures the non-local features while maintaining the real-time speed. Non-local features allow the tracker to be well-generalized  from one domain to another domain (e.g. from  ImageNet videos domain to  OTB/VOT videos domain).

Siamese architecture is widely used in face verification \cite{taigman2014closing}, face recognition \cite{schroff2015facenet},  One-shot recognition \cite{koch2015siamese} and object tracking \cite{bertinetto2016fully}. Most of Siamese-based trackers are built on AlexNet network \cite{krizhevsky2012imagenet} which is widely used in computer vision tasks. In Siamese-based trackers network, the input is a pair of images, the target image patch from the  very first frame and the search image patches which contain the  candidate objects from  later frames. The network learns a similarity function between the candidate patches and the target patch from the very first frame through the whole sequence. The output feature/response map is calculated based on the receptive field which is local and does not consider the object's context. The response map in such a case does not have  powerful expressive features to be robust to appearance changes.

In this paper, we present a novel network design for robust visual object tracking to improve the generalization capability of Siamese architecture. DensSiam network solves these issues and produces a non-local response map. DensSiam has two branches, the target branch that takes the input target patch from the first frame and the search branch that takes later images in the whole sequence. The target branch consists of dense blocks separated by a transition layer, each block has many convolutional layers and each transition layer has a convolutional layer with an average pool layer. Each dense layer is connected to every other layer in a feed-forward fashion.  The target response map is fed into the Self-Attention module to calculate response at a position as a weighted sum of the features at all positions. The search branch is the same architecture as the target branch except that it does not have the Self-Attention module since we calculate the similarity function between the target patch with non-local features and the candidate patches. Both target and search branches share the same network parameters across channels extracted from the same dense layers. To the best of our knowledge, this is the first densely Siamese network with a Self-Attention model.\\

To summarize, the main contributions of this work are three-fold.    
\begin{itemize}
   \item[$\bullet$] A novel end-to-end deep Densely-Siamese architecture is proposed for object tracking. The new architecture can capture the non-local features which are robust to appearance changes. Additionally, it reduces the number of shared parameters between layers while building up deeper network compared to other existing Siamese-based architectures commonly used in current state-of-the-art trackers.      
   \item[$\bullet$] An effective response map based on Self-Attention module that boosts the DensSiam tracker performance . The response map has no-local features and captures the semantic information about the target object.  
   \item[$\bullet$] The proposed architecture tackles the vanishing-gradient problem, leverages feature reuse and improves the generalization capability.  
   \end{itemize} 

The rest of the paper is organized as follows. We first introduce related work in Section \ref{Related work}. Section \ref{Proposed} details the proposed approach. We present the experimental results in Section \ref{Experiments}. Finally, Section \ref{Conclusions } concludes the paper.

\section{Related work}
\label{Related work}
There are extensive surveys on visual object tracking in the literatures \cite{li2013survey,salti2012adaptive,smeulders2014visual}. Recently, deep features have demonstrated breakthrough accuracies compared to handcrafted features. DeepSRDCF \cite{danelljan2015convolutional} uses deep features of a pretrained  CNN (e.g. VGG \cite{Karen}) from different layers and integrate them into a correlation filter.          Visual object tracking can be modeled as a similarity learning function in an offline  training phase. 
Siamese architecture consists of two branches, the target object branch and the search image branch. Siamese architecture takes the advantage of end-to-end learning. Consequently, the learned function can be evaluated online during tracking.  

The pioneering work for object tracking is the SiamFC \cite{bertinetto2016fully}. SiamFC has two branches, the target branch and the appearance branch, a correlation layer is used to calculate the correlation between the target patch and the candidate patches in the search image. The search image is  usually larger than the target patch to calculate the similarities  in a single evaluation. CFNet  \cite{valmadre2017end} improved SiamFC by introducing a differentiable correlation filter layer to the target branch to adapt the target model online. DSiam \cite{guo2017learning} uses a fast transfer motion to update the leaned model online. Significantly improved performance as it captures some information about the object's context.\\
SINT \cite{tao2016siamese} uses optical flow and formulates the visual object tracking as a verification problem within Siamese architecture, it has a better performance however, the speed dropped down from 86 to 4 frames per second. SA-Siam \cite{he2018twofold} uses two Siamese networks based on the original Siamese architecture \cite{bertinetto2016fully}. The first network for semantic information and the other one for appearance model. This architecture supranationally improved the tracker performance as it allows the semantic information of the appearance model representation to incorporate into the response map. However these trackers use the features taken directly from CNN which processes the information in a local neighborhood. Consequently the output features are local and do not have a lot information about the neighbourhood. Thus using convolutional layers alone is not effective to capture the generic appearance model.



\section{Proposed Approach}
\label{Proposed}

\begin{figure}
\centering
\includegraphics[width=\columnwidth]{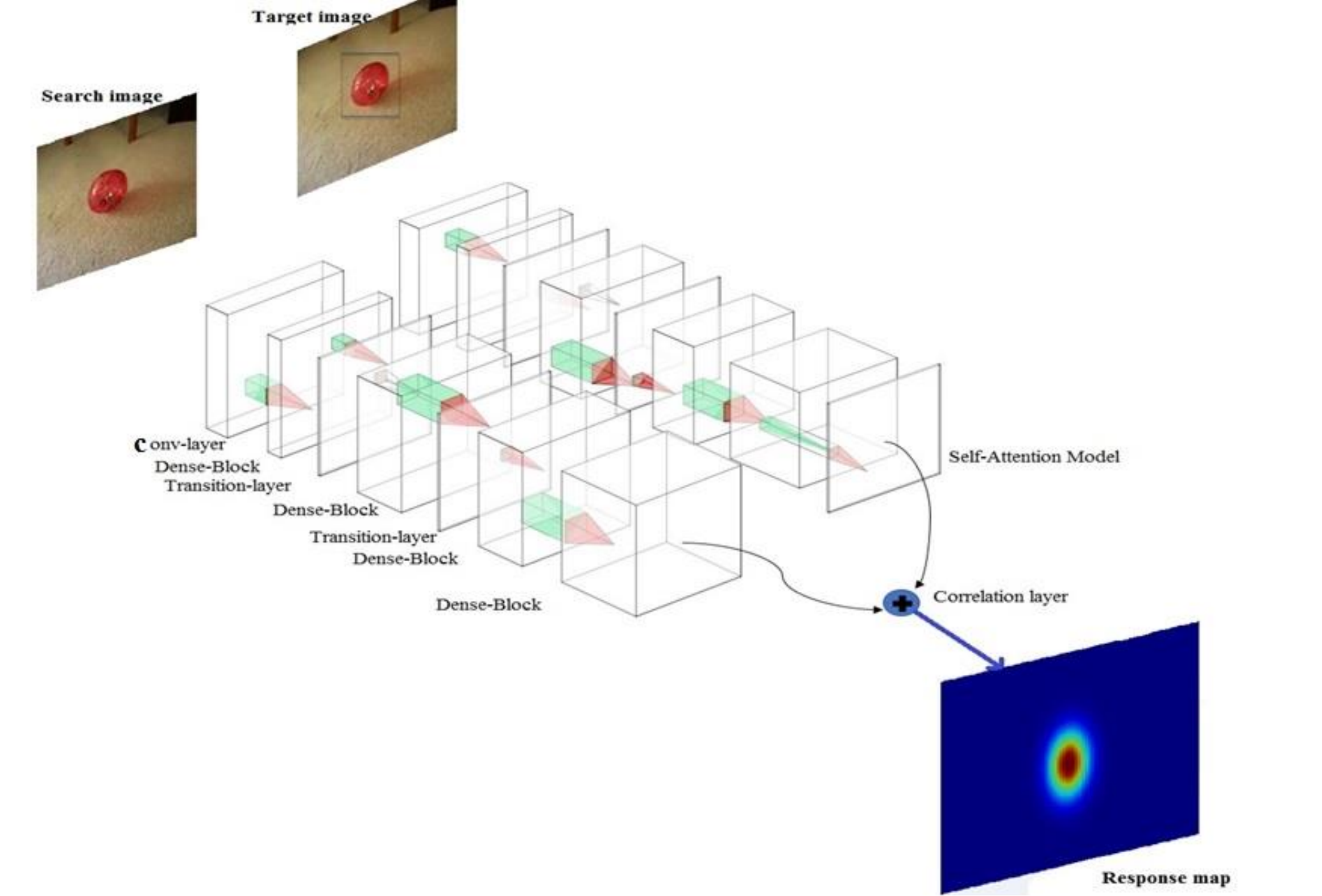}
\label{DensSiam}
\caption{The architecture of DensSiam tracker. The target branch has the Self-Attention model to capture the semantic information during offline training.}
\end{figure}

We propose effective and efficient deep architecture for visual tracking named Densely-Siamese network (DensSiam). Fig. \ref{DensSiam} shows the DensSiam architecture  of the proposed tracker. The core idea behind this design is that, using densely blocks separated by transition layers to build up Siamese network with Self-Attention model to capture non-local features. 

\subsection{Densely-Siamese Architecture}
\begin{figure*}[t!]
\centering
\includegraphics[width=.5\columnwidth]{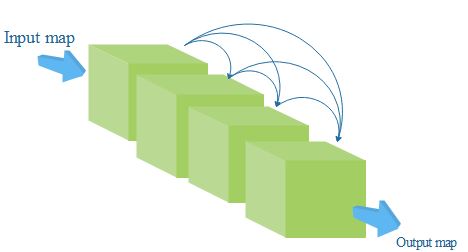}
\caption{ The internal structure of dense block without BN, ReLU, 1$\times$1 conv and Pooling layers shows the connections between convolutional layers in $Dense Block_2$}
\label{dense_block}
\end{figure*}

DensSiam architecture consists of two branches, the target branch and the appearance branch. The target branch architecture as follows: input-ConvNet-DenseBlock-TransitionLayer-DenseBlock-TransitionLayer-DenseBlock-DenseBlock-SelfAttention.\\
\textbf{Dense block}: Consists of  Batch Normalization (BN) \cite{Ioffe}, Rectified Linear Units (ReLU), pooling  and  Convolution layers, all dimensions are shown in Table \ref{tab1}. Each layer in dense block  takes all preceding feature maps as input and concatenate them. These connections  ensure that the network will   preserve the information needed from all preceding feature maps and  improve the information
flow between layers, and thus improves the generalization capability as shown in Fig. \ref{dense_block}. In traditional Siamese the output of $l^{th}$ layer is fed in as input to the $(l+1)^{th}$ layer we denote the output of $l^{th} $ layer as $x_l$. To formulate the information flow between layers in  dense block lets assume that the input tensor is $x_0 \in$ $\Re^{ C \times N \times D}$. Consequently, $l^{th}$ layer receives all feature maps according to this equation:

\begin{equation} \label {eq:x}
x_l= H_l ([x_0, x_1,...,x_{l-1}]),
\end{equation}   
Where $H_l$ is a three consecutive operations, batch normalization, ReLU and a 3 $\times$ 3 convolution and $[x_0, x_1,...,x_{l-1}]$ is a feature maps concatenation.\\
\textbf{Transition layer}: Consists of convolutional operation, average pooling and dropout \cite{srivastava2014dropout}. Adding dropout to dense block and transition layer  decreases the risk that DensSiam overfits to negative classes. 
DensSiam does not use padding operation since  it is a fully-convolutional and padding violates this property. Consequently, the size of feature maps in dense blocks varies and can not be concatenated. Therefore, transition layers are used to match up the size of dense blocks and concatenate them properly. 
\subsection{Self-Attention Model}
\begin{figure*}[t!]
\centering
\includegraphics[width=1\columnwidth]{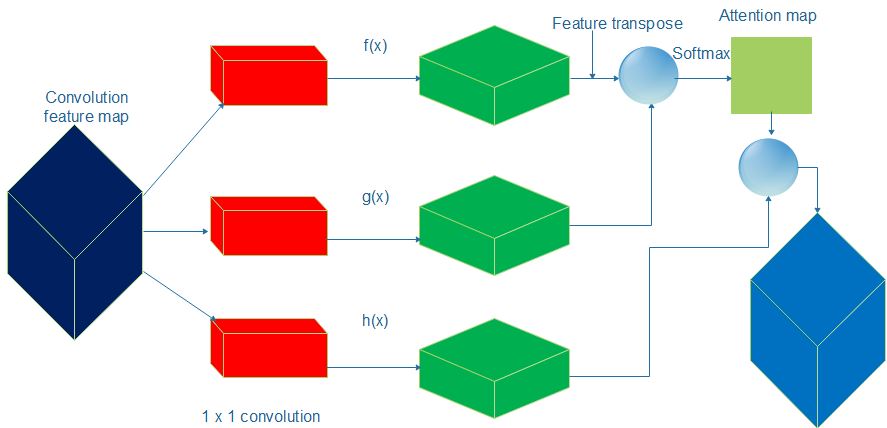}
\caption{ Self-Attention Model, the feature map is divided into three maps, blue circles are matrix multiplication. The input  and the output tensors are the same size.}
\label{Attention}
\end{figure*}

Attention mechanism was used in image classification \cite{wang2017residual}, multi-object tracking \cite {chu2017online}, pose estimation \cite{du2017rpan} and $etc$.
In addition to the new architecture of DensSiam and inspired by \cite{vaswani2017attention,NonLocal2018} that use the attention mechanism in object detection, DensSiam architecture integrates the Self-Attention model to target branch as shown in Fig.  \ref{Attention}.\\

Given the input tensor $x \in$ $\Re^{W \times H \times D}$ which is the output feature map of the target branch, Self-Attention model divides the feature map into three maps through  $1\times1$ convolutinal operation  to capture the attention, $f(x)$ and $g(x)$ can be calculated as follows:

\begin{equation} \label {eq:fx}
f(x)=W_f \times x,
\end{equation} 
\begin{equation} \label {eq:gx}
g(x)=W_g \times x,
\end{equation}
Where $W_f$, $W_g$ are the offline learned parameters. Attention feature map can be calculated as follows:
\begin{equation} \label {eq:phi}
\phi=\frac{\exp{(m)}}{\sum\limits_{i=1}^N\exp(m)},
\end{equation}
Where $\phi$ is the attention map weights and $m$= $f(x_i)^T \times g(x_i)$. 
Self-Attention feature map can be calculated as follows:
\begin{equation} \label {eq:att_map}
\sum\limits_{i=1}^N(\phi \times h(x_i)),
\end{equation}
Where $h(x) = W_h \times x$ and $W_h$ is the offline learned parameter. We use logistic loss for both dense block and Self-Attention model to calculate the weights using Stochastic Gradient Descent (SGD) as follows:

\begin{equation} \label {eq:loss}
l(y,v)= log(1+\exp(-yv)),
\end{equation}
Where $v$ is the single score value of target-candidate pair and $y \in {[-1 ,+1]}$ its ground truth label. To calculate the loss function for the feature map we use the mean of the loss over the whole map as follows:

\begin{equation} \label {eq:loss_mean}
L(y,v)= \frac {1}{N} \sum_{n\in N} l(y[n],v[n]),
\end{equation}
Finally, the search branch has the same architecture as target branch except the search branch does not have the Self-Attention model. The output of Self-Attention map and the output of the search branch are fed into correlation layer to learn the similarity function.

\section{Experimental Results}
\label{Experiments}
We divided the benchmarks to two sets, the validation set which includes OTB2013, OTB2015 and the testing set which includes VOT2015 VOT2016 and VOT2017. We provide the implementation details and hyper-parameters in the next subsection. 
\subsection{Implementation Details}
DensSiam  is pre-trained offline from scratch on the video object detection dataset of the ImageNet Large Scale Visual Recognition Challenge (ILSVRC15) \cite{russakovsky2015imagenet}. ILSVRC15 contains 1.3 million labelled frames in 4000 sequences and it has a wide variety of  objects which contribute to the generalization of the DensSiam network.\\ 
\textbf{Network architecture}. We adopt this architecture $input-ConvNet-DenseBlock_1-TransitionLayer-DenseBlock_2-TransitionLayer-DenseBlock_3-DenseBlock_4-SelfAttention$ as shown in Table   \ref{tab1}.\\
\textbf{Hyper-parameters settings}. Training is performed over 100 epochs, each with 53,200 sampled pairs. Stochastic gradient descent
(SGD) is applied with momentum of 0.9 to train the network. We adopt the mini-batches of size 8 and  the learning rate is annealed geometrically at each epoch from $10^{-3}$ to $10^{-8}$. We implement DensSiam  in TensorFlow \cite{abadi1603tensorflow} 1.8 framework. The experiments are performed on a PC with a Xeon E5 2.20 GHz CPU and a  Titan XP GPU. The testing speed of DensSiam is 60 fps.\\
\textbf{Tracking settings}. We adapt the scale variations by searching for the object on three scales $O^s$ where $O = 1.0375$ and $s= \{-2,0,2\}$. The input target image size is 127 $\times$ 127 and the search image size is $255 \times 255$. We use the linear interpolation to update the scale with a factor 0.764.

\begin{table*}[!t]

\centering
\caption{Data dimensions in DensSiam.}\label{tab1}
\begin{tabular}{c|c|c|c}
\hline
Layers & Output size  & Target branch  & Search branch \\
\specialrule{1.2pt}{0pt}{0pt}
 
$Convolution$ & Tensor (8, 61, 61, 72 )& 7 $\times$ 7 conv, stride 2
 &7 $\times$ 7 conv, stride 2 \\
\hline
\shortstack{$Dense Block_1$ \\ $ $}   & \shortstack{Tensor(8, 61, 61, 144) \\ $ $}   & \shortstack{\big[1 $\times $ 1 conv] $\times2$ \\ \big[3 $\times$3  conv] $\times 2$} & \shortstack{\big[1 $\times $ 1 conv] $\times2$ \\ \big[3 $\times$3  conv] $\times 2$} \\
\hline
$Transition Layer$ &Tensor (8, 30, 30, 36)   & 1 $\times$ 1 conv , average pool & 1 $\times$ 1 conv , average pool\\
\hline
\shortstack{$Dense Block_2$ \\ $ $} & \shortstack{Tensor(8, 30, 30, 180) \\ $ $}   & \shortstack{\big[1 $\times $ 1 conv] $\times4$ \\ \big[3 $\times$3  conv] $\times 4$}&\shortstack{\big[1 $\times $ 1 conv] $\times4$ \\ \big[3 $\times$3  conv] $\times 4$}\\
\hline
$Transition Layer$ &Tensor(8, 15, 15, 36)  & 1 $\times$ 1 conv , average pool & 1 $\times$ 1 conv , average pool\\
\hline
\shortstack{$Dense Block_3$ \\ $ $} &\shortstack{ Tensor (8, 15, 15, 252) \\$ $ }  & \shortstack{\big[1 $\times $ 1 conv] $\times 6$ \\ \big[3 $\times$3  conv] $\times 6$} &\shortstack{\big[1 $\times $ 1 conv] $\times6$ \\ \big[3 $\times$3  conv] $\times 6$}\\
\hline
$Dense Block_4$ & Tensor(8, 9, 9, 128) &\big [7 $\times$ 7 conv] $\times$ 3  &\big [7 $\times$ 7 conv] $\times$ 3 \\
\hline
$Self-Attention$ & Tensor(8, 9, 9, 128) & \big[1 $\times$1  conv] $\times 3$ &-- \\
\hline
\end{tabular}
\end{table*}

%
%

%

\begin{table*}[!t]
    \begin{minipage}{.5\textwidth}
      \caption{Comparison with the state-of-the-art trackers including the top four non-realime for VOT2015.}\label{VOT2015}
\begin{tabular}{l|c|c|c|c}
\specialrule{1.2pt}{0pt}{0pt}
Tracker &  A & R & EAO  & FPS\\
\specialrule{1.2pt}{0pt}{0pt}
MDNet     & 0.60 & 0.69 & 0.38 & 1 \\
DeepSRDCF &  0.56 & 1.05 & 0.32 & $<$ 1\\

EBT & 0.47 & 1.02 &0.31& 4.4 \\

SRDCF & 0.56 & 1.24& 0.2 & 5 \\
\hline
BACF &  0.59& 1.56& $-$ & 35 \\
EAST &  0.57&1.03& 0.34& 159 \\
Staple & 0.57& 1.39&0.30 &80 \\
SamFC & 0.55& 1.58& 0.29& 86 \\
\hline
\textbf{DensSiam(ours)}  & 0.619 & 1.24& 0.34 & 60\\
\specialrule{1.2pt}{0pt}{0pt}
\end{tabular}

  \end{minipage}
  \begin{minipage}{.5\textwidth}
      \caption{Comparison with the state-of-the-art trackers on VOT2016.}\label{VOT2016}
\begin{tabular}{l|c|c|c|c}
\specialrule{1.2pt}{0pt}{0pt}
Tracker &  A & R & EAO  & FPS\\
\specialrule{1.2pt}{0pt}{0pt}
ECOhc  & 0.54 & 1.19 & 0.3221& 60 \\
Staple & 0.54 & 1.42& 0.2952 &80\\
STAPLE+ & 0.55 & 1.31& 0.2862&$>$ 25\\ 
SiamRN & 0.55&1.36& 0.2766& $>$ 25 \\
GCF& 0.51&1.57 &0.2179& $>$ 25\\
\hline
\textbf{DensSiam(ours)} & 0.56& 1.08 & 0.3310& 60\\

\specialrule{1.2pt}{0pt}{0pt}
\end{tabular}
     
    \end{minipage}
\end{table*}

\begin{table*}[!t]
\centering
\caption{Comparison with the state-of-the-art trackers on VOT2017.}\label{VOT2017}
\begin{tabular}{l|c|c|c|r}
\specialrule{1.2pt}{0pt}{0pt}
Tracker &  A & R & EAO  & FPS\\
\specialrule{1.2pt}{0pt}{0pt}
SiamDCF & 0.500 &0.473& 0.249&60\\
ECOhc& 0.494&0.435&0.238&60\\
CSRDCF++&0.453&0.370&0.229& $>$ 25\\
SiamFC & 0.502& 0.585&0.188& 86\\
SAPKLTF& 0.482&0.581&0.184& $>$ 25 \\
Staple & 0.530& 0.688& 0.169& $>$ 80 \\
ASMS & 0.494& 0.623 & 0.169& $>$ 25\\
\hline
\textbf{DensSiam(ours)} & 0.540& 0.350& 0.250 &60 \\ 

\specialrule{1.2pt}{0pt}{0pt}
\end{tabular}
\end{table*}

\subsection{Comparison with the State-of-the-Arts}
In this section we use VOT  toolkit standard metrics \cite{kristan2016visual}, accuracy (A), robustness (R) and expected average overlap (EAO).
Table \ref{VOT2015} shows the comparison of DensSaim with MDNet \cite{nam2016learning}, DeepSRDCF \cite{danelljan2015convolutional}, EBT \cite{zhu2016beyond}, BACF \cite{galoogahi2017learning}, EAST \cite{huang2017learning}, Staple \cite{bertinetto2016staple} and SiamFC \cite{bertinetto2016fully}. The four top trackers are non-realtime  trackers and we still outperform them. In terms of accuracy, DensSiam is about 2 \% higher than MDNet while it is the best second after MDNet in terms of expected average overlap. DensSiam is the highest in terms of robustness score in real-time trackers. We also report the resutls of our tracker on VOT2016 and VOT2017 as shown in Table \ref{VOT2016} and Table \ref{VOT2017}. The comparison includes ECOhc \cite{danelljan2017eco}, SiamFC \cite{bertinetto2016fully}, SiamDCF \cite{wang2017dcfnet}, Staple \cite{bertinetto2016staple}, CSRDCF++ \cite{lukezic2017discriminative}. DensSiam  outperforms all trackers in terms of accuracy, robustness score and expected average overlap.

\section{Conclusions and Future Work}
This paper proposed DensSiam, a new Siamese architecture for object tracking. DensSiam uses non-local features to represent the appearance model in such a way that allows the deep feature map to be robust to appearance changes. DenSaim allows different feature levels (e.g. low level and high level features) to flow through the network layers without vanishing gradients and improves the generalization capability . The resulting tracker greatly benefits from the Densely-Siamese architecture with Self-Attention model and substantially increases the accuracy and robustness while decreasing the number of shared network parameters. The architecture of DensSiam can be extended to other tasks of computer vision such as object verification, recognition and detection since it is general Siamese framework.    
\label{Conclusions }


\bibliographystyle{splncs04}

\bibliography{References}

\begin{thebibliography}{10}
\providecommand{\url}[1]{\texttt{#1}}
\providecommand{\urlprefix}{URL }
\providecommand{\doi}[1]{https://doi.org/#1}

\bibitem{abadi1603tensorflow}
Abadi, M., Agarwal, A., Barham, P., Brevdo, E., Chen, Z., Citro, C., Corrado,
  G., Davis, A., Dean, J., Devin, M., et~al.: Tensorflow: large-scale machine
  learning on heterogeneous distributed systems. arxiv preprint (2016). arXiv
  preprint arXiv:1603.04467

\bibitem{alahari2015thermal}
Alahari, K., Berg, A., Hager, G., Ahlberg, J., Kristan, M., Matas, J.,
  Leonardis, A., Cehovin, L., Fernandez, G., Vojir, T., et~al.: The thermal
  infrared visual object tracking vot-tir2015 challenge results. In: Computer
  Vision Workshop (ICCVW), 2015 IEEE International Conference on. pp. 639--651.
  IEEE (2015)

\bibitem{bertinetto2016staple}
Bertinetto, L., Valmadre, J., Golodetz, S., Miksik, O., Torr, P.H.: Staple:
  Complementary learners for real-time tracking. In: Proceedings of the IEEE
  conference on computer vision and pattern recognition. pp. 1401--1409 (2016)

\bibitem{bertinetto2016fully}
Bertinetto, L., Valmadre, J., Henriques, J.F., Vedaldi, A., Torr, P.H.:
  Fully-convolutional siamese networks for object tracking. In: European
  conference on computer vision. pp. 850--865. Springer (2016)

\bibitem{chen2015deepdriving}
Chen, C., Seff, A., Kornhauser, A., Xiao, J.: Deepdriving: Learning affordance
  for direct perception in autonomous driving. In: Computer Vision (ICCV), 2015
  IEEE International Conference on. pp. 2722--2730. IEEE (2015)

\bibitem{chu2017online}
Chu, Q., Ouyang, W., Li, H., Wang, X., Liu, B., Yu, N.: Online multi-object
  tracking using cnn-based single object tracker with spatial-temporal
  attention mechanism. In: 2017 IEEE International Conference on Computer
  Vision (ICCV).(Oct 2017). pp. 4846--4855 (2017)

\bibitem{danelljan2017eco}
Danelljan, M., Bhat, G., Khan, F.S., Felsberg, M.: Eco: Efficient convolution
  operators for tracking. In: Proceedings of the 2017 IEEE Conference on
  Computer Vision and Pattern Recognition (CVPR), Honolulu, HI, USA. pp. 21--26
  (2017)

\bibitem{danelljan2015convolutional}
Danelljan, M., Hager, G., Shahbaz~Khan, F., Felsberg, M.: Convolutional
  features for correlation filter based visual tracking. In: Proceedings of the
  IEEE International Conference on Computer Vision Workshops. pp. 58--66 (2015)

\bibitem{danelljan2015learning}
Danelljan, M., Hager, G., Shahbaz~Khan, F., Felsberg, M.: Learning spatially
  regularized correlation filters for visual tracking. In: Proceedings of the
  IEEE International Conference on Computer Vision. pp. 4310--4318 (2015)

\bibitem{du2017rpan}
Du, W., Wang, Y., Qiao, Y.: Rpan: An end-to-end recurrent pose-attention
  network for action recognition in videos. In: IEEE International Conference
  on Computer Vision. vol.~2 (2017)

\bibitem{galoogahi2017learning}
Galoogahi, H.K., Fagg, A., Lucey, S.: Learning background-aware correlation
  filters for visual tracking. In: Proceedings of the 2017 IEEE Conference on
  Computer Vision and Pattern Recognition (CVPR), Honolulu, HI, USA. pp. 21--26
  (2017)

\bibitem{guo2017learning}
Guo, Q., Feng, W., Zhou, C., Huang, R., Wan, L., Wang, S.: Learning dynamic
  siamese network for visual object tracking. In: Proc. IEEE Int. Conf. Comput.
  Vis. pp.~1--9 (2017)

\bibitem{he2018twofold}
He, A., Luo, C., Tian, X., Zeng, W.: A twofold siamese network for real-time
  object tracking. In: Proceedings of the IEEE Conference on Computer Vision
  and Pattern Recognition. pp. 4834--4843 (2018)

\bibitem{henriques2015high}
Henriques, J.F., Caseiro, R., Martins, P., Batista, J.: High-speed tracking
  with kernelized correlation filters. IEEE Transactions on Pattern Analysis
  and Machine Intelligence  \textbf{37}(3),  583--596 (2015)

\bibitem{huang2017learning}
Huang, C., Lucey, S., Ramanan, D.: Learning policies for adaptive tracking with
  deep feature cascades. In: IEEE Int. Conf. on Computer Vision (ICCV). pp.
  105--114 (2017)

\bibitem{Ioffe}
Ioffe, S., Szegedy, C.: Batch normalization: Accelerating deep network training
  by reducing internal covariate shift. In: ICML. vol.~37, pp. 448--456 (2015)

\bibitem{kendall2015posenet}
Kendall, A., Grimes, M., Cipolla, R.: Posenet: A convolutional network for
  real-time 6-dof camera relocalization. In: Computer Vision (ICCV), 2015 IEEE
  International Conference on. pp. 2938--2946. IEEE (2015)

\bibitem{koch2015siamese}
Koch, G., Zemel, R., Salakhutdinov, R.: Siamese neural networks for one-shot
  image recognition. In: ICML Deep Learning Workshop. vol.~2 (2015)

\bibitem{kristan2016visual}
Kristan, M., Matas, J., Leonardis, A., Felsberg, M., Cehovin, L., Fernandez,
  G., Vojir, T., Hager, G., Nebehay, G., Pflugfelder, R.: The visual object
  tracking vot2016 challenge results. In: European Conference on Computer
  Vision Workshop (2016)

\bibitem{krizhevsky2012imagenet}
Krizhevsky, A., Sutskever, I., Hinton, G.E.: Imagenet classification with deep
  convolutional neural networks. In: Advances in neural information processing
  systems. pp. 1097--1105 (2012)

\bibitem{lenc15understanding}
Lenc, K., Vedaldi, A.: Understanding image representations by measuring their
  equivariance and equivalence. In: Proceedings of the {IEEE} Conf. on Computer
  Vision and Pattern Recognition ({CVPR}) (2015)

\bibitem{li2013survey}
Li, X., Hu, W., Shen, C., Zhang, Z., Dick, A., Hengel, A.V.D.: A survey of
  appearance models in visual object tracking. ACM transactions on Intelligent
  Systems and Technology (TIST)  \textbf{4}(4), ~58 (2013)

\bibitem{li2014scale}
Li, Y., Zhu, J.: A scale adaptive kernel correlation filter tracker with
  feature integration. In: European Conference on Computer Vision. pp.
  254--265. Springer (2014)

\bibitem{lukezic2017discriminative}
Lukezic, A., Vojir, T., Zajc, L.C., Matas, J., Kristan, M.: Discriminative
  correlation filter with channel and spatial reliability. In: Proceedings of
  the IEEE Conference on Computer Vision and Pattern Recognition. vol.~2 (2017)

\bibitem{ma2015hierarchical}
Ma, C., Huang, J.B., Yang, X., Yang, M.H.: Hierarchical convolutional features
  for visual tracking. In: Proceedings of the IEEE International Conference on
  Computer Vision. pp. 3074--3082 (2015)

\bibitem{molchanov2016online}
Molchanov, P., Yang, X., Gupta, S., Kim, K., Tyree, S., Kautz, J.: Online
  detection and classification of dynamic hand gestures with recurrent 3d
  convolutional neural network. In: Proceedings of the IEEE Conference on
  Computer Vision and Pattern Recognition. pp. 4207--4215 (2016)

\bibitem{mueller2017context}
Mueller, M., Smith, N., Ghanem, B.: Context-aware correlation filter tracking.
  In: Proc. IEEE Conf. Comput. Vis. Pattern Recognit.(CVPR). pp. 1396--1404
  (2017)

\bibitem{nam2016learning}
Nam, H., Han, B.: Learning multi-domain convolutional neural networks for
  visual tracking. In: Proceedings of the IEEE Conference on Computer Vision
  and Pattern Recognition. pp. 4293--4302 (2016)

\bibitem{russakovsky2015imagenet}
Russakovsky, O., Deng, J., Su, H., Krause, J., Satheesh, S., Ma, S., Huang, Z.,
  Karpathy, A., Khosla, A., Bernstein, M., et~al.: Imagenet large scale visual
  recognition challenge. International Journal of Computer Vision
  \textbf{115}(3),  211--252 (2015)

\bibitem{salti2012adaptive}
Salti, S., Cavallaro, A., Di~Stefano, L.: Adaptive appearance modeling for
  video tracking: Survey and evaluation. IEEE Transactions on Image Processing
  \textbf{21}(10),  4334--4348 (2012)

\bibitem{schroff2015facenet}
Schroff, F., Kalenichenko, D., Philbin, J.: Facenet: A unified embedding for
  face recognition and clustering. In: Proceedings of the IEEE conference on
  computer vision and pattern recognition. pp. 815--823 (2015)

\bibitem{Karen}
Simonyan, K., Zisserman, A.: Very deep convolutional networks for large-scale
  image recognition. In: International Conference on Learning Representations
  (2015)

\bibitem{smeulders2014visual}
Smeulders, A.W., Chu, D.M., Cucchiara, R., Calderara, S., Dehghan, A., Shah,
  M.: Visual tracking: An experimental survey. IEEE transactions on pattern
  analysis and machine intelligence  \textbf{36}(7),  1442--1468 (2014)

\bibitem{srivastava2014dropout}
Srivastava, N., Hinton, G., Krizhevsky, A., Sutskever, I., Salakhutdinov, R.:
  Dropout: A simple way to prevent neural networks from overfitting. The
  Journal of Machine Learning Research  \textbf{15}(1),  1929--1958 (2014)

\bibitem{taigman2014closing}
Taigman, Y., Yang, M., Ranzato, M., Wolf, L.: Closing the gap to human-level
  performance in face verification. deepface. In: IEEE Computer Vision and
  Pattern Recognition (CVPR) (2014)

\bibitem{tao2016siamese}
Tao, R., Gavves, E., Smeulders, A.W.: Siamese instance search for tracking. In:
  Computer Vision and Pattern Recognition (CVPR), 2016 IEEE Conference on. pp.
  1420--1429. IEEE (2016)

\bibitem{valmadre2017end}
Valmadre, J., Bertinetto, L., Henriques, J., Vedaldi, A., Torr, P.H.:
  End-to-end representation learning for correlation filter based tracking. In:
  Computer Vision and Pattern Recognition (CVPR), 2017 IEEE Conference on. pp.
  5000--5008. IEEE (2017)

\bibitem{vaswani2017attention}
Vaswani, A., Shazeer, N., Parmar, N., Uszkoreit, J., Jones, L., Gomez, A.N.,
  Kaiser, {\L}., Polosukhin, I.: Attention is all you need. In: Advances in
  Neural Information Processing Systems. pp. 6000--6010 (2017)

\bibitem{wang2017residual}
Wang, F., Jiang, M., Qian, C., Yang, S., Li, C., Zhang, H., Wang, X., Tang, X.:
  Residual attention network for image classification. arXiv preprint
  arXiv:1704.06904  (2017)

\bibitem{wang2017dcfnet}
Wang, Q., Gao, J., Xing, J., Zhang, M., Hu, W.: Dcfnet: Discriminant
  correlation filters network for visual tracking. arXiv preprint
  arXiv:1704.04057  (2017)

\bibitem{NonLocal2018}
Wang, X., Girshick, R., Gupta, A., He, K.: Non-local neural networks. CVPR
  (2018)

\bibitem{wu2015object}
Wu, Y., Lim, J., Yang, M.H.: Object tracking benchmark. IEEE Transactions on
  Pattern Analysis and Machine Intelligence  \textbf{37}(9),  1834--1848 (2015)

\bibitem{zhu2016beyond}
Zhu, G., Porikli, F., Li, H.: Beyond local search: Tracking objects everywhere
  with instance-specific proposals. In: Proceedings of the IEEE Conference on
  Computer Vision and Pattern Recognition. pp. 943--951 (2016)

\end{thebibliography}

\end{document}